\documentclass[10pt,twocolumn,letterpaper]{article}

\usepackage{abc}
\usepackage{times}
\usepackage{epsfig}
\usepackage{graphicx}
\usepackage{subfigure}
\usepackage{amsmath}
\usepackage{amssymb}
\usepackage{url}
\usepackage[section]{placeins}

\abcfinalcopy

\ifabcfinal\fi
\begin{document}

\title{SeesawFaceNets: sparse and robust face verification model for mobile platform}

\author{Jintao.Zhang \\
	DiDiChuxing\\
	\tt\small jtzhangcas@gmail.com
}

\maketitle
\thispagestyle{empty}

\begin{abstract}
Deep Convolutional Neural Network (DCNNs) come to be the most widely used solution for most computer vision related tasks, and one of the most important application scenes is face verification. Due to its high-accuracy performance, deep face verification models of which the inference stage occurs on cloud platform through internet plays the key role on most prectical scenes. However, two critical issues exist: First, individual privacy may not be well protected since they have to upload their personal photo and other private information to the online cloud backend. Secondly, either training or inference stage is time-comsuming and the latency may affect customer experience, especially when the internet link speed is not so stable or in remote areas where mobile reception is not so good, but also in cities where building and other construction may block mobile signals. 
Therefore, designing lightweight networks with low memory requirement and computational cost is one of the most practical solutions for face verification on mobile platform.
In this paper, a novel mobile network named SeesawFaceNets, a simple but effective model, is proposed for productively deploying face recognition for mobile devices.
Dense experimental results have shown that our proposed model SeesawFaceNets outperforms the baseline MobilefaceNets, with only {\bf66\%}(146M VS 221M MAdds) computational cost, smaller batch size and less training steps, and SeesawFaceNets achieve comparable performance with other SOTA model e.g. mobiface with only {\bf54.2\%}(1.3M VS 2.4M) parameters and {\bf31.6\%}(146M VS 462M MAdds) computational cost, It is also eventually competitive against large-scale deep-networks face recognition on all 5 listed public validation datasets, with {\bf6.5\%}(4.2M VS 65M) parameters and {\bf4.35\%}(526M VS 12G MAdds) computational cost.
\end{abstract}

\section{Introduction}

Deep Convolutional Neural Network (DCNNs) have revolutionized traditional machine learning applications with unprecedented performace as well as strong robustness, the performance of DCNNs exceeds human-level accuracy in more and more vision or non-vision related tasks. Most of the computer vision applications, e.g. image classification \cite{Kaiming-Resnet,Luu_FR_CVPR2016,Luu_FR_CVPR2015}, object detection \cite{girshick2014rcnn,Luu_FD_CVPR2018,Luu_FD_FG2018,Luu_FR_ICCV2017,Zheng_FD_BTAS2016,Zhu_FD_CVPRW2016,Le_FD_ICPR2016}
image segmentation \cite{he2017maskrcnn,Long-FCN,Luu_Seg_PR2018,Luu_Seg_TIP2018,le_miccai_2018}, and face modeling \cite{Duong_IJCV2018} etc., deploy DCNNs to achieve the state-of-the-art. However, using deeper neural network with hundreds of layer and millions of parameters comes at heavy cost. Due to the high computational cost and the fact that still many operations are less optimised in existing deep learning frameworks on different hardware platforms\cite{hu2018squeeze}, DCNNs require high computational resources beyond the capabilities of daily mobile and embedded applications.

To design such powerful neural networks, powerful GPU clusters are required so as to provide sufficient GPU memory and reduce training/inference time. Due to this hardware limitation, accordingly, there are several new advances in compressing DCNNs. Generally, most of these works utilize Mimic Networks \cite{hinton2015distilling, Li2017MimickingVE,Wei2018QuantizationMT}, Network Pruning \cite{Han-LBWs,Han-Deep,Liu2017LearningEC}, Efficient build block designing \cite{iandola2016squeezenet, Howard-MobileNet,zhang2018shufflenet,sandler2018mobilenetv2,sun2019IGCV3,zhang2019SeesawNet,ma2018shufflenet,tan2019mnasnet, howard2019searching}, Binary Networks \cite{Itay-BNN,Matt-BinaryNet,Rastegari-xnor,CourbariauxBD15}. These previously proposed methods improve the efficiency of the training and inference stage of the model with only small loss in accuracy. However, since face verification tasks are expected to be robust such as the very deep features of millions of facial subjects discriminated \cite{Duong2018mobiface}, speedup face recognition model with above compressed networks is still challenging task.

The main contributions of this work is a novel sparse and robust deep neural network which greatly minimizes the conputational and memory required for either training or inference stage while retaining the accuracy for mobile face verification task. Compared to previous work, we present our contributions in SeesawFaceNets approach as follows: (1) we improve the baseline face verification model MobileFaceNets \cite{chen2018MobileFaceNets} by utilizing the Seesaw-block and SE-block to achieve further lighter-weight model but higher accuracy performance. (2) the proposed method SeesawFaceNets is then applied in face verification and trained an end-to-end on face verification database. In addition, we extensively experiment our proposed SeesawFaceNets on mobile-based network on face verification tasks with alternative state-of-the-art face recognition databases including LFW, AgeDB-30, CFP-FP, CPLFW and CALFW, and the results show that our SeesawFaceNets obtains stable improved efficiency compared with state-of-the-art lightweight and mobile CNNs for face verification with less computational cost, memory requirement and training steps.

\section{Prior Work}

There has been a rising interest in lightweight neural architectures designing to obtain an optimal balance between accuracy and efficiency during the past several years. Most of previous lightweight models could be categorized into efficient designed networks and pruned networks, binarized networks, quantized networks. Our proposed work focus on efficient neural networks designing for face verification tasks on mobile platforms.

Face recognition is one of the most important tasks in computer vision related area.
To reduce the huge memory requirement and computational cost of classical large face verification model, MobileFaceNets \cite{chen2018MobileFaceNets} were presented mainly based on MobileNetV2 framework. Dense face verification benchmark experiments on alternative datasets in \cite{chen2018MobileFaceNets} show that MobileFaceNets achieve significantly superior accuracy as well as speedup on mobile devices even through it contains obviously fewer parameters and computational cost than mobile networks which directly utilize image classfication backbones\cite{Howard-MobileNet,zhang2018shufflenet,sandler2018mobilenetv2}. MobiFace \cite{Duong2018mobiface} introduces efficient memory and low cost operators by adopting fast downsampling\cite{qin2018fd} and bottleneck residual block \cite{sandler2018mobilenetv2}, and achieves 99.7\% on LFW database and 91.3\% on Megaface database. ShrinkTeaNet \cite{duong2019shrinkteanet} presents a teacher-student learning paradigm to train a portable student network which achieves 99.77\% on LFW, 95.14\% on CFP-FP and 95.64\% on Megaface protocols. AirFace \cite{li2019airface} proposes a novel loss function and improved the the perfomance of MobileFaceNets with deeper and wider network and attention module to achieve better result on face verification tasks. However, still there are much work to be done to effectively run deep learning framework on variable mobile devices which will be expected to show diverse performances \cite{ignatov2018ai}.

\section{Our Proposed SeesawFaceNets}

This section Firstly introduces network design strategies we deployed in this work to construct sparse but robust model. Then, by adopting these strategies, the SeesawFaceNets architecture for face verification on mobile devices is proposed. Meanwhile, we proposed a practical issue caused by the huge GPU requirement when training large neural networks, dense further experiments results prove that our proposed SeesawFaceNets outperforms other previous mobile networks, and our deeper and wider version model - DW-SeesawFaceNet could achieve comparable performance with state of the art large networks with much less computational cost and memory requirement on face verification tasks.

\subsection{Network Design Strategy}
\textbf{Seesaw block}. Seesaw block is introduced in \cite{zhang2019SeesawNet}. however, to deploy Seesaw block for face verification tasks, we modified the original Seesaw block and add non-linearity after the second pointwise convolution layer as MobileNetV2 and MobilefaceNets does.
There are 2 key factors of this type of block compared with MobilefaceNets: (1) Seesaw block utilizes uneven pointwise group convolution; (2) channel permute/shuffle operation is applied to enable the information flow between uneven convolution groups. 
We use swish \cite{ramachandran2017searching, howard2019searching} as the non-linearity, which is slightly better for face verification than using PReLU\cite{ramachandran2017searching}.

The residual connection is employed in a bottleneck block whose input channel number equals to the output channel number. This type of build blocks is shown to have the capabilities of preventing manifold collapse during transformation and also enable the network to represent
more complex functions \cite{sandler2018mobilenetv2}. The use of Seesaw block is introduced in Seesaw-Net\cite{zhang2019SeesawNet} where the original pointwise convolutions within Bottleneck Residual block are replaced with uneven goup convolutions and channel permute/shuffle operation is deployed to enable information flow between convolutional groups. With Seesaw-block, Seesaw-Net achieve notable better performance and efficiency beyond mobilenet-v2 and IGCV3 on dense classification datasets. Moreover, additional ablation experiments was included in \cite{zhang2019SeesawNet} to verify adaptability on small expansion ratio, which implies Seesaw block may has the potential to construct lighter-weight but efficient model for mobile tasks. 

Based on mobilenet-v2\cite{sandler2018mobilenetv2}, mobilenet-v3\cite{howard2019searching} introduce 3 main updates and achieve new SOTA on several mobile classification, detection and segmentation tasks: 1. Apply the squeeze and excite in the residual layer; 2. Use different nonlinearity depending on the layer; 3. Deploy NAS specifically on number of input/output channels for each basic blocks. The main perpose of our paper is to boost performance as well as efficiency of the baseline model-mobilefacenets, so we just adopt the SE-block and different nonlinearity to mobilefacenets and replace the mobilenet-v2 block with Seesaw-shuffle/Seesaw-share block which we expected will save notable computation cost.

\textbf{Seesaw-shuffle and Seesaw-share block}. With limited computational resource on mobile platform, Seesaw-Net\cite{zhang2019SeesawNet} introduce 2 types of build block named Seesaw-shuffle and Seesaw-share block so as to enable information flow between uneven convolutional groups in each stage. As \cite{zhang2019SeesawNet} mentioned, only one channel permute/shuffle operation is included in Seesaw-shuffle block and it could be avoided with low-level language implementation instead of calling high-level API of deep learning frameworks. For high-level deep learning API development, Seesaw-share block introduce channel share operation that will enables the information flow between convolutional groups like channel shuffle operation without any channel permute/shuffle operation, which implies Seesaw-share block could achieve better efficiency compared with channel permute/shuffle based block under the limited computational budgets, since memory transfer lead to more time and power consumption on mobile platform. Generally, high efficiency light-model could be constructed with Seesaw-shuffle and Seesaw-share block. 

\begin{figure}[htbp]
\centering
\subfigure[Seesaw-shuffle block]{
\begin{minipage}[t]{0.4\linewidth}
\centering
\includegraphics[width=1.3in]{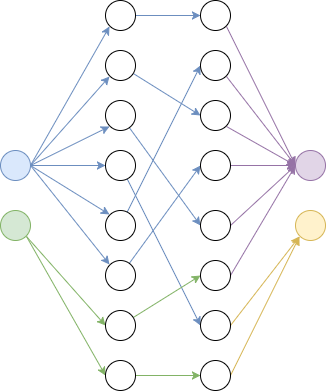}
\end{minipage}
}%
\subfigure[Seesaw-share block]{
\begin{minipage}[t]{0.4\linewidth}
\centering
\includegraphics[width=1.3in]{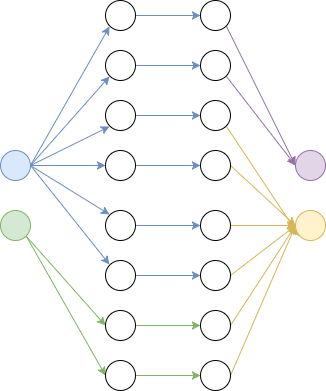}
\end{minipage}
}%
\centering
\caption{ The structure of the proposed Seesaw-shuffle and Seesaw-share block}
\end{figure}

\begin{table}[htbp]
	\small
	\centering
	\caption{Model Architecture of SeesawFaceNet for facial feature embedding. $/2$ means stride of the operator is 2. ``RBlock'' and ``Block'' respectively indicate the Seesaw Block with and without Residual learning connection.}
	\label{tb:SeesawFaceNet_structure} 
	\begin{tabular}{|c|l|}
		\hline
		\textbf{Input} & \textbf{Operator} \\  \hline \hline
		112 $\times$ 112 $\times$ 3 & 3 $\times$ 3 Conv, $/2$, 64\\
		\hline
		56 $\times$ 56 $\times$ 64 & 3 $\times$ 3 DWconv, 64\\
		\hline
		56 $\times$ 56 $\times$ 64 &  Block  1 $\times$ 
		                      \Bigg\{
                              \begin{tabular}{lll}
                              1 $\times$ 1 Conv, 128 \\
                              3 $\times$ 3 DWconv, $/2$, 128 \\
                              1 $\times$ 1 Conv, Linear, 64 
                              \end{tabular}
                             \\
		\hline
		28 $\times$ 28 $\times$ 64 & RBlock 4 $\times$
		                      \Bigg\{
                              \begin{tabular}{lll}
                              1 $\times$ 1 Conv, 128 \\
                              3 $\times$ 3 DWconv, 128 \\
                              1 $\times$ 1 Conv, Linear, 64 
                              \end{tabular}
                            \\
        \hline
		28 $\times$ 28 $\times$ 64 & Block  1 $\times$
		                      \Bigg\{
                              \begin{tabular}{lll}
                              1 $\times$ 1 Conv, 256 \\
                              3 $\times$ 3 DWconv, $/2$, 256 \\
                              1 $\times$ 1 Conv, Linear, 128 
                              \end{tabular}
                            \\
        \hline
		14 $\times$ 14 $\times$ 128 & RBlock 6 $\times$
		                      \Bigg\{
                              \begin{tabular}{lll}
                              1 $\times$ 1 Conv, 256 \\
                              3 $\times$ 3 DWconv, 256 \\
                              1 $\times$ 1 Conv, Linear, 128 
                              \end{tabular}
                            \\
        \hline
		14 $\times$ 14 $\times$ 128 & Block 1 $\times$
		                      \Bigg\{
                              \begin{tabular}{lll}
                              1 $\times$ 1 Conv, 512 \\
                              3 $\times$ 3 DWconv, $/2$, 512 \\
                              1 $\times$ 1 Conv, Linear, 128 
                              \end{tabular}
                            \\
        \hline
		7 $\times$ 7 $\times$ 128 & RBlock 2 $\times$
		                      \Bigg\{
                              \begin{tabular}{lll}
                              1 $\times$ 1 Conv, 256 \\
                              3 $\times$ 3 DWconv, 256 \\
                              1 $\times$ 1 Conv, Linear, 128 
                              \end{tabular}
                            \\
        \hline
        7 $\times$ 7 $\times$ 128 & 1 $\times$ 1 Conv, 512\\
       \hline
       7 $\times$ 7 $\times$ 512 & linear GD7 $\times$ 7 Conv, 512\\
        \hline
        1 $\times$ 1 $\times$ 512 & linear 1 $\times$ 1 Conv\\
        \hline
	\end{tabular}
\end{table}

\begin{table}[htbp]
	\small
	\centering
	\caption{Model Architecture of SeesawFaceNet which adopt the hyper parameters of MobiFace for facial feature embedding. $/2$ means stride of the operator is 2. ``RBlock'' and ``Block'' respectively indicate the Seesaw Block with and without Residual connection.}
	\label{tb:SeesawFaceNet_mobi_structure} 
	\begin{tabular}{|c|l|}
		\hline
		\textbf{Input} & \textbf{Operator} \\  \hline \hline
		112 $\times$ 112 $\times$ 3 & 3 $\times$ 3 Conv, $/2$, 64\\
		\hline
		56 $\times$ 56 $\times$ 64 & 3 $\times$ 3 DWconv, 64\\
		\hline
		56 $\times$ 56 $\times$ 64 &  Block  1 $\times$ 
		                      \Bigg\{
                              \begin{tabular}{lll}
                              1 $\times$ 1 Conv, 128 \\
                              3 $\times$ 3 DWconv, $/2$, 128 \\
                              1 $\times$ 1 Conv, Linear, 64 
                              \end{tabular}
                             \\
		\hline
		28 $\times$ 28 $\times$ 64 & RBlock 2 $\times$
		                      \Bigg\{
                              \begin{tabular}{lll}
                              1 $\times$ 1 Conv, 128 \\
                              3 $\times$ 3 DWconv, 128 \\
                              1 $\times$ 1 Conv, Linear, 64 
                              \end{tabular}
                            \\
        \hline
		28 $\times$ 28 $\times$ 64 & Block  1 $\times$
		                      \Bigg\{
                              \begin{tabular}{lll}
                              1 $\times$ 1 Conv, 256 \\
                              3 $\times$ 3 DWconv, $/2$, 256 \\
                              1 $\times$ 1 Conv, Linear, 128 
                              \end{tabular}
                            \\
        \hline
		14 $\times$ 14 $\times$ 128 & RBlock 3 $\times$
		                      \Bigg\{
                              \begin{tabular}{lll}
                              1 $\times$ 1 Conv, 256 \\
                              3 $\times$ 3 DWconv, 256 \\
                              1 $\times$ 1 Conv, Linear, 128 
                              \end{tabular}
                            \\
        \hline
		14 $\times$ 14 $\times$ 128 & Block 1 $\times$
		                      \Bigg\{
                              \begin{tabular}{lll}
                              1 $\times$ 1 Conv, 512 \\
                              3 $\times$ 3 DWconv, $/2$, 512 \\
                              1 $\times$ 1 Conv, Linear, 128 
                              \end{tabular}
                            \\
        \hline
		7 $\times$ 7 $\times$ 128 & RBlock 6 $\times$
		                      \Bigg\{
                              \begin{tabular}{lll}
                              1 $\times$ 1 Conv, 256 \\
                              3 $\times$ 3 DWconv, 256 \\
                              1 $\times$ 1 Conv, Linear, 128 
                              \end{tabular}
                            \\
        \hline
        7 $\times$ 7 $\times$ 128 & 1 $\times$ 1 Conv, 512\\
       \hline
       7 $\times$ 7 $\times$ 512 & linear GD7 $\times$ 7 Conv, 512\\
        \hline
        1 $\times$ 1 $\times$ 512 & linear 1 $\times$ 1 Conv\\
        \hline
	\end{tabular}
\end{table}

\subsection{SeesawFaceNets}
In this section, we propose a novel model - SeesawFaceNets , sparse and robust face verification. 
Given an $112 \times 112 \times 3$ RGB face image, SeesawFaceNets aims at efficient facial feature embedding with low computational cost and memory requirement.
Inspired by the strategies presented in previous sections, Seesaw block is adopt as the basic build block of SeesawFaceNets, compared to MobileFaceNets which utilize the original inverted linear bottlenecks as the basic build block. Table \ref{tb:SeesawFaceNet_structure} represents the architecture of SeesawFaceNets.

With the squeeze-and-excitation optimization\cite{hu2018squeeze}, \cite{deng2019arcface,sandler2018mobilenetv2,howard2019searching} achieves slightly better performance on face recognition and image classification tasks. Seesaw-Net\cite{zhang2019SeesawNet} achieves better performance on mobile classification tasks compared with previous work but no experiments with squeeze-and-excitation optimization included since it focus on basic build block design. We adopt squeeze-and-excitation optimization in our proposed SeesawFaceNets, since we expect the squeeze-and-excitation optimization will improves channel interdependencies of Seesaw-block with limited extra computational cost.

In our implementation, Swish is used for the non-linear activation function due to its accuracy improvement over PReLU and ReLU function.
The embedding size during our work is fixed to 512 rather than 128 in \cite{chen2018MobileFaceNets}, since better performance could be obtained by 512 with only few amount of computational cost introduced.

\subsection{Batch Size and Learning Rate}

Recent deep neural network training is typically based on mini-batch stochastic gradient optimization.  An milestone work on relation between batch size and learning rate is the linear scaling rule adopted by Krizhevsky \cite{krizhevsky2014one}, which reported a $1\%$ increase of error on the validation set of ILSVRC 2012 when increasing the minibatch size from 128 to 1024. \cite{goyal2017accurate} show how to maintain accuracy across a much broader regime of minibatch sizes while also present an informal discussion of the linear scaling rule and why it may be effective.

In our experiments, we adopt similar learning schedule from the baseline model-mobilefacenets except the batch size due to limited GPU memory. However, we hypothesize the above linear scaling rule will not effect result obvously across a much narrow regime of minibatch sizes compared to \cite{krizhevsky2014one} and \cite{goyal2017accurate}, and we found that the model converges very slowly if we set 0.025 as the initial learning rate which may require longer training epochs. We set dense control experiments which adopt same batch size for all models to make fair comparisons. The experiment result also verified our view, our proposed SeesawFaceNets outperforms mobilefacenets and MobiFace with similar training schedule and achieve state of the art on mobile face recognition tasks.

\section{Experiments}
\subsection{Datasets}

\begin{table}[htbp]
\begin{center}
\begin{tabular}{c|c|c}
\hline
 Datasets   & \#Identity & \#Image\\
\hline
MS1MV2  & 85K  & 5.8M \\   
\hline
LFW \cite{huang2008labeled}   & 5,749  & 13,233 \\
CFP-FP \cite{sengupta2016frontal} & 500 & 7,000\\
AgeDB-30 \cite{Moschoglou2017AgeDB} & 568 & 16,488 \\
CPLFW \cite{zheng2018cross} & 5,749  & 11,652 \\
CALFW \cite{zheng2017cross} & 5,749  & 12,174 \\
\hline
\end{tabular}
\end{center}
\caption{Face datasets for training and testing.}
\label{table:datasets}
\vspace{-4mm}
\end{table}

\textbf{MS1MV2}\cite{guo2016ms}  is introduced as a semi-automatic refined version of the MS-Celeb-1M dataset. As given in Table 1, we employ MS1MV2 as our training data in order to conduct fair comparison with other methods. During training, we explore public face verification database (\eg LFW \cite{huang2008labeled}, CFP-FP \cite{sengupta2016frontal}, AgeDB-30 \cite{Moschoglou2017AgeDB}) to check the improvement from different settings. Besides the most widely used LFW \cite{huang2008labeled} dataset, we also report the performance of SeesawFaceNets on the recent large-pose and large-age datasets(\eg CPLFW \cite{zheng2018cross} and CALFW \cite{zheng2017cross}) to further verify robustness of the models.
This training data has no overlapping with the testing data.

\begin{table*}[!]
\begin{center}
\begin{tabular}{c|c|c|c|c|c|c|c|c}
\hline
Method   &\#Image &Training epoch & Params &MAdds & Batch size & LFW & CFP-FP & AgeDB-30 \\
\hline
Mobiface \cite{Duong2018mobiface}       & 3.8M & 1024 & 2.4M & 462M &1024 & {\bf 99.70} & - & -\\
\hline
LMobileNetE\cite{deng2019arcface}         & 3.8M & 27 & 26.7M& - & 512 & 99.50 &  88.50 &  96.06\\
MobileFaceNet \cite{chen2018MobileFaceNets}       & 0.5M & 8 & 1.0M & 221M &512 & 99.28 & - & 93.05\\
MobileFaceNet \cite{chen2018MobileFaceNets}       & 3.8M & 8 & 1.0M & 221M &512 & 99.55 & - & 96.07\\
MobileFaceNet(*)      & 5.8M & 16 & 1.2M & 221M &160 & 99.52 & 91.64 & 95.92\\
MobileFaceNet(*)      & 5.8M & 16 & 1.2M & 221M &192 & 99.57 & 91.39 & 96.03\\
MobileFaceNet(*)      & 5.8M & 16 & 1.2M & 221M &256 & 99.55 & 91.46 & 96.03\\
MobileFaceNet(*)      & 5.8M & 24 & 1.2M & 221M &256 & 99.57 & 91.81 & 96.32\\
Seesaw-shareFaceNet   & 5.8M & 16 & 1.3M & 146M &160 & 99.55 & {\bf 93.09} & {\bf 96.42} \\
Seesaw-shuffleFaceNet  & 5.8M & 16 & 1.3M & 146M &160 & {\bf 99.70} & 92.03 & 96.30 \\
Seesaw-shuffleFaceNet  & 5.8M & 16 & 1.3M & 146M &192 & {\bf 99.60} & {\bf92.90} & {\bf96.85} \\
Seesaw-shuffleFaceNet(mobi)  & 5.8M & 16 & 2.8M & 154M &224 & {\bf 99.65} & {\bf93.07} & {\bf96.48} \\
\hline
\end{tabular}
\end{center}
\vspace{-2mm}
\caption{Verification results ($\%$) of different model. MobileFaceNet(*) stands for our implement of MobileFaceNet with 512D embedding size(Training dataset of MobileFaceNet\cite{chen2018MobileFaceNets}: CASIA-Webface/MS-Celeb-1M(cleaned), LMobileNetE: MS-Celeb-1M(refined); Our implement of MobileFaceNet/Seesaw-shuffleFaceNet/Seesaw-shareFaceNet/Seesaw-shuffleFaceNet(mobi): MS1MV2. For each validation dataset, top-3 accuracy records in bold)}
\label{table:light model result}
\vspace{-3mm}
\end{table*}

\begin{table*}[!]
\begin{center}
\begin{tabular}{c|c|c|c|c|c|c|c|c|c|c}
\hline
Method &Training epoch & Params &MAdds & Batch size & LFW & CFP-FP & AgeDB-30 & CPLFW & CALFW \\
\hline
HUMAN-Individual\cite{deng2019arcface}  & - & -& - & - & 97.27 &  - &  - & 81.21 &  82.32\\
HUMAN-Fusion\cite{deng2019arcface}  & - & -& - & - & 99.85 &  - &  - & 85.24 &  86.50\\
\hline
Arcface\cite{deng2019arcface}  & 16 & 65M& 12.1G & 512 & 99.83 &  98.37 &  98.15 & 92.08 &  95.45\\
AirFace\cite{li2019airface} & 61 & - & 1G & 1024 & 99.267 &  94.514 &  - & - &  -\\
ShrinkTeaNet-MFNR\cite{duong2019shrinkteanet} & - & 3.73M& - & 512 & 99.77 &  95.14 &  - & - &  -\\
DW-SeesawFaceNet V1 & 16 & 4.1M & 514M &128 &  99.80 &  96.39 &  97.52 & 91.77 &  96.07\\
DW-SeesawFaceNet V2 & 16 & 4.2M & 526M &128 &  99.80 &  97.24 &  97.55 & 91.98 &  95.98\\
\hline
\end{tabular}
\end{center}
\vspace{-2mm}
\caption{Verification results ($\%$) of different model. Training dataset: CASIA-Webface for AirFace and MS1MV2 for other model. For AirFace, we list the best evaluation results with different loss function}
\label{table:large model result}
\vspace{-3mm}
\end{table*}

The comparisons between different models in terms of both accuracy/model size/MAdds are represented. Here we only count MAdds of the convolutional and fullly connected layers, and sigmoid operations not take into account. An key difference between sigmoid/swish and other non-linearity functions is that sigmoid/swish belongs to transcendental functions. To the best of our knowledge, mainstream deep learning hardware will include dedicated processing unit for transcendental functions implement. e.g. NVIDIA GPU since fermi series, Google TPU include activation pipeline, dedicated ASICs, table-lookup and CORDIC(coordinate rotation digital computer) based implement on FPGA platform\cite{zhang2013anagpufpga}.

\subsection{Implementation Details}
During the face preprocessing stage, MTCNN\cite{zhang2016joint} face detection and alignment algorithm is applied to detect all faces and their five landmark points.
Then, with the returned face landmark points, each face is aligned and cropped into a template with the size of $112 \times 112 \times 3$. The template is normalized into $[-1,1]$ by subtracting the mean pixel value, i.e. 127.5, and then divided by 128.

For training, we adopt Stochastic Gradient Descent (SGD) as the optimizer with the batch size of 164/192/256 due to limited GPU memory. It is also a common but key practical issue for DCNNs training stage that lightweight model could be trained with relative large batch size with same GPU drvices compared to large model. The momentum parameter is set to $0.9$, the initial learning rate is set to $0.1$ and decreases by a factor of 10 periodically at 9, 13, 15 epochs for 16-epochs training schedule. All of the proposed SeesawFaceNets models in this work are trained from scratch by ArcFace loss with a angular margin m=0.5 for a fair performance comparison with other state of the art models.

\subsection{Verification Results}

According to dense experiments on various public validation dataset, our proposed SeesawFaceNets outperform LMobileNetE and MobileFaceNet on all listed Verification datasets: SeesawFaceNets always performs better than MobileFaceNet with 66\% of the MAdds operation even with less batch size and training steps, and Seesaw-shareFaceNet does not include extra channel permute/shuffle operations which may introduce cost when implement with high-level language or mainstream deep learning frameworks\cite{zhang2019SeesawNet}.

\section{Ablation Study}

\subsection{Adaptability on other SOTA mobile networks on face verification tasks}

To further verify the performance of proposed improved seesaw block in face recognition tasks, we adopt the hyper-parameters and network structure of MobiFace to construct a Seesaw-shuffleFaceNet(mobi)(see Table \ref{tb:SeesawFaceNet_mobi_structure}). 

According to the verification result, the Seesaw-shuffleFaceNet(mobi) could also achieve better performance on all listed benchmark dataset than MobileFaceNet with 69.7\%(154M VS 221M MAdds), and  achieve comparable performance with SOTA model-mobiface on LFW dataset with only 54.2\%(1.3M VS 2.4M) parameters and 31.6\%(146M VS 462M MAdds) computational cost.

\subsection{Adaptability on deeper and wider networks}
In order to pursue ultimate performance, we construct deeper and wider version of Seesaw-shuffleFaceNet named DW-SeesawFaceNet and train the novel model with same training schedule and hyper parameters but less batch size due to limited available GPU memory, which also implies that our DW-SeesawFaceNet could be trained from scratch by mini-batch size of 128 with only 22GB GPU memory requirement with pytorch 1.0\cite{paszke2017automatic}.

Inverted residual bottleneck is a baseline structure of dense SOTA mobile (\cite{sandler2018mobilenetv2, sun2019IGCV3, zhang2019SeesawNet, tan2019mnasnet, howard2019searching}) first introduced in \cite{sandler2018mobilenetv2} since improved information flow enabled by shortcut connections in residual structure, however, when the input channel number does not equal to the output channel, shortcut connections is removed in Inverted residual bottleneck structure. To strengthen the feature reuse, firstly we add a skip connection branch comprises a 2*2 max pooling and a 1*1 convolution layers to all Inverted bottleneck blocks without residual structure whose filter stride is 2, we named this version of deeper and wider networks(see Table \ref{tb:DW-SeesawFaceNet_structure}) as DW-SeesawFaceNet V2 and named the baseline version without the skip connection branch as DW-SeesawFaceNet V1. The performance on listed public benchmark datasets show that the second version gives better result.

According to the verification result, the DW-SeesawFaceNet V1 and V2 could reach comparable performance on all listed public Verification datasets, with 6.5\% parameters size and 4.35\% computational cost compared to large-scale deep face recognition network\cite{deng2019arcface}.

\begin{table}[htbp]
	\small
	\centering
	\caption{Model Architecture of DW-SeesawFaceNet for facial feature embedding. $/2$ means stride of the operator is 2. ``RBlock'' and ``Block'' respectively indicate the Seesaw Block with and without Residual connection.}
	\label{tb:DW-SeesawFaceNet_structure} 
	\begin{tabular}{|c|l|}
		\hline
		\textbf{Input} & \textbf{Operator} \\  \hline \hline
		112 $\times$ 112 $\times$ 3 & 3 $\times$ 3 Conv, $/2$, 96\\
		\hline
		56 $\times$ 56 $\times$ 96 & 3 $\times$ 3 DWconv, 96\\
		\hline
		56 $\times$ 56 $\times$ 96 &  Block  1 $\times$ 
		                      \Bigg\{
                              \begin{tabular}{lll}
                              1 $\times$ 1 Conv, 128 \\
                              3 $\times$ 3 DWconv, $/2$, 128 \\
                              1 $\times$ 1 Conv, Linear, 96 
                              \end{tabular}
                             \\
		\hline
		28 $\times$ 28 $\times$ 96 & RBlock 8 $\times$
		                      \Bigg\{
                              \begin{tabular}{lll}
                              1 $\times$ 1 Conv, 192 \\
                              3 $\times$ 3 DWconv, 192 \\
                              1 $\times$ 1 Conv, Linear, 96 
                              \end{tabular}
                            \\
        \hline
		28 $\times$ 28 $\times$ 96 & Block  1 $\times$
		                      \Bigg\{
                              \begin{tabular}{lll}
                              1 $\times$ 1 Conv, 384 \\
                              3 $\times$ 3 DWconv, $/2$, 384 \\
                              1 $\times$ 1 Conv, Linear, 192 
                              \end{tabular}
                            \\
        \hline
		14 $\times$ 14 $\times$ 192 & RBlock 12 $\times$
		                      \Bigg\{
                              \begin{tabular}{lll}
                              1 $\times$ 1 Conv, 384 \\
                              3 $\times$ 3 DWconv, 384 \\
                              1 $\times$ 1 Conv, Linear, 192 
                              \end{tabular}
                            \\
        \hline
		14 $\times$ 14 $\times$ 192 & Block 1 $\times$
		                      \Bigg\{
                              \begin{tabular}{lll}
                              1 $\times$ 1 Conv, 768 \\
                              3 $\times$ 3 DWconv, $/2$, 768 \\
                              1 $\times$ 1 Conv, Linear, 192 
                              \end{tabular}
                            \\
        \hline
		7 $\times$ 7 $\times$ 192 & RBlock 4 $\times$
		                      \Bigg\{
                              \begin{tabular}{lll}
                              1 $\times$ 1 Conv, 384 \\
                              3 $\times$ 3 DWconv, 384 \\
                              1 $\times$ 1 Conv, Linear, 192 
                              \end{tabular}
                            \\
        \hline
        7 $\times$ 7 $\times$ 192 & 1 $\times$ 1 Conv, 512\\
       \hline
       7 $\times$ 7 $\times$ 512 & linear GD7 $\times$ 7 Conv, 512\\
        \hline
        1 $\times$ 1 $\times$ 512 & linear 1 $\times$ 1 Conv\\
        \hline
	\end{tabular}
\end{table}

\section{Conclusion}
Inspired by different network design strategies, this paper has further presented a novel simple but high-performance deep network for face recognition, named SeesawFaceNets. Experiments on dense public face verification benchmarks have shown the efficiency of our SeesawFaceNets in terms of both accuracy and computation cost. Although the model is very small, its performance on both testing benchmarks is competitive against other large-scale deep face recognition network.

\section{Acknowledgments}
We would like to thank Xiangyu Zhang for his helpful suggestion and Chi Nhan Duong team for their feedback.

{\small
\bibliographystyle{ieee}
\bibliography{submission_example}
}

\end{document}